\documentclass[journal]{IEEEtran}

\usepackage{amsmath,amssymb}
\usepackage{graphicx}
\usepackage{booktabs}
\usepackage{multirow}
\usepackage{cite}
\usepackage{stfloats}
\usepackage{array}
\usepackage{caption}
\usepackage{threeparttable}
\usepackage[hidelinks,breaklinks=true]{hyperref}

\title{ZAPS-DA: Zero-Phase Action Policy Smoothing with
Decoupled Actor for Continuous Control in Reinforcement Learning}

\author{Faiq~Shamass
\thanks{F.~Shamass is an independent researcher, San Diego, CA, USA
(e-mail: faiq.shamass@gmail.com).}
}

\begin{document}
\maketitle

\begin{abstract}
Continuous control policies trained with off-policy reinforcement learning frequently
exhibit high-frequency action jitter, rendering direct deployment on physical actuators
impractical. Post-hoc filtering attenuates jitter but introduces phase lag; embedding
smoothness penalties in the actor's loss couples them with the RL gradient and conflates
reward regression with over-aggressive smoothing. We present
\textbf{ZAPS-DA}, a framework that reduces action jitter at deployment with negligible
phase lag and no post-processing. ZAPS-DA pairs an unmodified main actor (trained by the base
RL loss) with a separate decoupled actor trained via supervised imitation of zero-phase 
filtered targets stored in the replay buffer. The deployed policy is the
decoupled actor: a feed-forward map from the current observation to a smooth action,
with no inference-time filter and no action-history input --- a mechanism we term
\emph{causal distillation of a non-causal filter}. A magnitude-matched MSE loss provides
zero-hyperparameter portability across optimizer classes. Validated with Soft Actor-Critic
and a Savitzky--Golay filter in two driving simulators using paired $n{=}150$ evaluation
protocols: on MetaDrive (anchor protocol),
ZAPS-DA reduces steering jitter by $14$--$21\times$ and throttle jitter by $3$--$5\times$
(all $p<10^{-4}$, Bonferroni-corrected) while matching task-completion ($p=0.28$ success,
$p=0.31$ crash) at a $6.3\%$ reward cost; on a custom Webots adaptive cruise control
environment, the same SG configuration produces a Pareto improvement --- reward parity
($p=0.121$), $8$--$45\times$ steering jitter reduction, and total task-failure rate
reduced from $2.0\%$ to $0.7\%$. Against CAPS, the standard penalty-based smoothing
baseline --- evaluated at both its auto-entropy and its native fixed-entropy operating
points, with its penalty weight, spatial noise, and entropy coefficient each re-tuned to
the environment --- ZAPS-DA reaches $14.7\times$ steering-jitter reduction versus CAPS's
best $3.2\times$ at matched seeds and protocol, a $\approx 4.6\times$ gap attained with no
per-environment tuning of the smoothness signal and with post-hoc applicability to
already-trained policies.
\end{abstract}


\section{Introduction}

\IEEEPARstart{R}{einforcement} learning has demonstrated remarkable success in continuous
control tasks, yet deploying trained policies on physical systems frequently reveals a
critical practical limitation: action jitter. Policies trained with off-policy algorithms
such as SAC produce temporally inconsistent action sequences --- consecutive outputs that
oscillate rapidly around a mean value --- despite converging to high task reward. In
autonomous vehicle control, this manifests as bang-bang speed commands and high-frequency
steering oscillations that cause mechanical wear on actuators, degrade passenger comfort,
and render direct deployment on real hardware impractical.

The root cause is structural. Off-policy algorithms with feedforward critics evaluate each
state--action pair independently, with no temporal-consistency constraint between
consecutive actions. In vehicle control, plant inertia acts as a mechanical low-pass filter
on actuation commands, so smooth and jittery commands of the same average magnitude produce
nearly identical next-state distributions. The critic sees no Q-gradient distinguishing
them, and the policy is free to oscillate. We term this the \emph{flat Q-landscape problem}:
a structural blind spot in feedforward critics that makes action jitter invisible to the
learning signal. Post-hoc filtering attenuates jitter but introduces phase lag proportional
to the filter window --- unacceptable in safety-critical scenarios. Zero-phase filters such
as Savitzky--Golay~\cite{sg_filter} eliminate phase lag through symmetric non-causal
processing, but require future samples unavailable at deployment.

We present \textbf{ZAPS-DA} (Zero-phase Action Policy Smoothing with Decoupled Actor), a
framework that resolves this tension by confining the non-causal operation to training
time. ZAPS-DA combines two insights: first, while a zero-phase filter cannot be applied
causally at deployment, it can be applied non-causally during training --- where future
actions are available in the replay buffer's history window; second, the smoothness signal
need not, and should not, share a gradient with the task objective. Initial experiments
adding a Lagrangian-tuned smoothness penalty directly to the actor loss either saturated
the dual variable without producing smooth behavior or destabilised policy
learning~\cite{pid_lag} --- symptomatic of a structural property of any formulation that
couples smoothness with the actor's RL gradient (\S\ref{sec:related}), motivating an
architectural separation. ZAPS-DA pairs an unmodified
main actor (trained by the standard SAC loss) with a separate decoupled actor trained via
supervised imitation of zero-phase-filtered targets stored in the replay buffer. The
decoupled actor is the deployed artifact --- a feed-forward policy with no inference-time
filter, no action-history input, and no consultation of the main actor. We term this
mechanism \emph{causal distillation of a non-causal filter}.

The central novelty is \emph{architectural} rather than in the choice of smoothing
operator: prior action-smoothness methods add a penalty term to the actor's own loss,
coupling the smoothness gradient with the RL objective; ZAPS-DA instead moves the smoothness
signal out of the main actor entirely and into a separately-trained decoupled actor whose
gradient never reaches the policy --- a separation that also uniquely enables the method to
be applied post-hoc to an already-trained policy. The contributions of this work are:
\begin{itemize}
\item \textbf{The ZAPS-DA framework} --- a dual-actor training-time architecture that
decouples smoothness and reward objectives, with a zero-phase filter applied non-causally
to action histories in the replay buffer as a supervised target. The deployed decoupled
actor is feed-forward with no post-processing and negligible phase lag.
\item \textbf{Magnitude-matched MSE loss} --- a cumulative running-mean ratio that scales
the imitation loss to match actor-loss magnitude, providing zero-hyperparameter portability
across adaptive and non-adaptive optimizers.
\item \textbf{A controlled comparison against CAPS}~\cite{caps}, the standard penalty-based
action-smoothing baseline, on matched environment, seeds, and evaluation protocol, with CAPS
evaluated at both its auto-entropy and its native fixed-entropy operating points: ZAPS-DA
reaches $14.7\times$ steering-jitter reduction where CAPS attains at most $3.2\times$ ---
even after per-environment re-tuning of its penalty weight, spatial noise, and entropy
coefficient --- a $\approx 4.6\times$ gap that isolates the benefit of architectural
decoupling from the choice of smoothness oracle (\S\ref{sec:caps}).
\item \textbf{Empirical validation across two simulators} --- on MetaDrive: $14$--$21\times$
steering jitter reduction at task-completion parity and $6.3\%$ reward cost; on Webots ACC
with the same SG configuration: a Pareto improvement (reward parity, $8$--$45\times$ steering jitter
reduction, task-failure rate $2.0\% \to 0.7\%$).
\end{itemize}

\section{Related Work}
\label{sec:related}

\textbf{Action smoothness in RL.} CAPS~\cite{caps} is the canonical approach: it adds
auxiliary penalty terms --- on consecutive-output differences (temporal) and on output
sensitivity to nearby states (spatial) --- directly to the policy loss with tunable
coefficients $\lambda_T, \lambda_S$. This design has two structural consequences that
motivate ours. First, the smoothness signal lives \emph{inside} the actor's RL gradient and
is local (a $1$-step temporal comparison). Second, because reward and smoothness gradients
share the actor's parameters, the penalty must be present during training and cannot be
applied to an already-trained policy. ZAPS-DA differs on both counts: its smoothness signal
is non-causal (a zero-phase filter over a centered action-history window stored in the
replay buffer) and is supervised through a \emph{separate} actor whose gradient never
reaches the main policy. We compare directly against CAPS under matched conditions
(\S\ref{sec:caps}): architectural decoupling lets ZAPS-DA reach a $\approx 4.6\times$
greater jitter reduction than CAPS attains at its best native operating point, and lets
the smoother be applied post-hoc.

\textbf{Knowledge distillation.} DAgger~\cite{dagger} and GAIL~\cite{gail} distill
behavioral teachers into students~\cite{distill} --- the teacher encodes task knowledge
from experience or demonstration. ZAPS-DA instead distills a \emph{signal-processing
oracle}: a zero-phase filter encoding only temporal smoothness properties, requiring no
pre-trained expert and adding no task knowledge the policy did not already possess.

\textbf{The merged-objective confound.} The natural midpoint between CAPS and ZAPS-DA keeps
a single actor but replaces the CAPS penalty with an SG-imitation MSE penalty under a
dual-tuned $\lambda$~\cite{pid_lag}; in our experiments this either saturated the dual
variable without producing smooth behavior or destabilised policy learning (Appendix~\ref{app:lagrangian}: bounded $\lambda$ pinned at its ceiling, main-actor steering JE $\approx 22\times$
above ZAPS-DA's). The failure mode is structural to \emph{any} merged-objective formulation
--- dual-tuned or fixed-weight, SG-MSE or CAPS-style adjacency penalty: when smoothness and
reward gradients share an actor's parameters, an observed reward regression cannot be
cleanly attributed to suboptimal RL hyperparameters versus over-aggressive smoothing.
ZAPS-DA's architectural decoupling sidesteps this confound by construction and, as a
by-product, lets the framework be applied to previously trained policies --- post-hoc
smoothness without retraining, which no penalty-based method affords.

\textbf{Temporal consistency and post-filtering.} Action-conditioned~\cite{andrychowicz}
and recurrent~\cite{drqn} policies introduce temporal dependencies through architecture;
post-hoc low-pass or moving-average filters~\cite{robotics_textbook} attenuate jitter but
introduce phase lag proportional to the filter window. We quantify this trade-off
(\S\ref{sec:caps}): a causal filter matched to ZAPS-DA's steering-jitter reduction incurs
$\approx 0.84$\,s of phase lag, whereas ZAPS-DA's deployed decoupled actor is feed-forward,
requires no action-history input, and emits a smoothed action in one forward pass at
negligible lag.

\section{Methodology}

\textbf{Preliminaries.} Our main actor uses Soft Actor-Critic~\cite{sac} unmodified ---
same loss, same hyperparameters, same twin critic, same $\tanh$-squashed Gaussian policy.
Given matched seeds and hyperparameters, the main actor in any ZAPS-DA training run is
bit-for-bit identical to a standalone vanilla SAC run, a property we exploit when reporting
baseline comparisons (\S\ref{sec:results}). The Savitzky--Golay (SG) filter~\cite{sg_filter}
is the zero-phase filter we use as the smoothness oracle: it fits a degree-$p$ polynomial
to a centered window of $w$ samples by least squares and evaluates the fit at the center
(zero-phase by symmetry), yielding one dot product per output with closed-form symmetric
coefficients. We use $p=2$ and odd $w \geq 5$ throughout, with windows specified per action
dimension to match each signal's frequency content. Two SG properties matter: (i) the
centered fit can slightly overshoot the input range near sharp transitions, requiring an
output clamp; (ii) longer windows have more aggressive negative side-lobes, an effect
visible in the window ablation (\S\ref{sec:ablations}).

\subsection{Framework Overview}

ZAPS-DA pairs a standard off-policy actor $\pi_\phi$ --- trained by the unmodified base
RL algorithm --- with a secondary feed-forward decoupled actor $\pi_\psi$ that shares
the same input space (current observation $s$) and $\tanh$-squashed Gaussian output
structure. The two actors share neither parameters nor gradients. The replay buffer is
augmented to store, alongside each transition $(s_t, a_t, r_t, s_{t+1}, d_t)$, a
centered zero-phase-filtered version of the action at time $t$, denoted $\tilde{a}_t$.
At each gradient step the main actor, twin critics, and entropy coefficient $\alpha$ are
updated by the unmodified base-RL losses on $(s_t, a_t, r_t, s_{t+1}, d_t)$, while the
decoupled actor is updated by a magnitude-matched supervised MSE between its squashed
output and $\tilde{a}_t$ (\S\ref{sec:loss}). The decoupled actor never receives a
policy-gradient update; its gradient flows only through the supervised MSE. The deployed
policy is $\pi_\psi$ --- a single feed-forward map $s_t \mapsto \hat{a}_t$ with no
inference-time filter, no main-actor consultation, and no action-history input.

Generating $\tilde{a}_t$ requires a non-causal computation: the SG filter at time $t$
reads samples at times $t{+}1,\ldots,t{+}w/2$. ZAPS-DA confines this to a \emph{history
buffer} maintained inside the environment wrapper, holding the last $N$ transitions in
chronological order. Once the buffer is full, the SG filter is applied to the centered
window and the middle-slot transition --- for which all forward-side samples now exist
--- is pushed to the replay buffer with its filtered target attached. The current
transition is held in the history buffer until it itself rolls to the middle slot.
Diagram~1 illustrates this data flow.

\begin{figure}[!b]
  \centering
  \vspace{-1ex}
  \includegraphics[width=0.85\columnwidth]{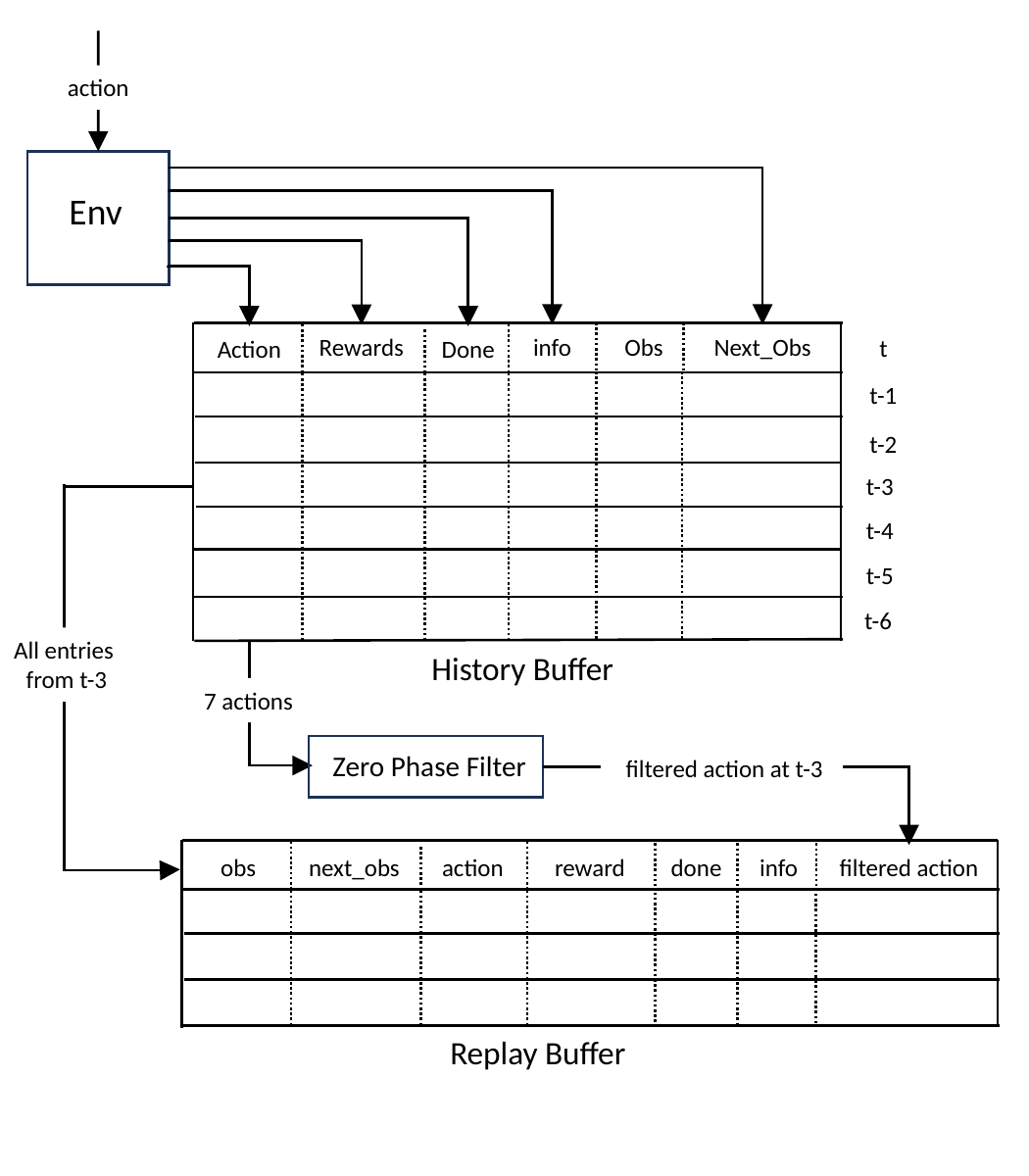}
  \vspace{-1ex}
  \caption*{\textbf{Diagram 1.} ZAPS-DA training-time data flow. The environment wrapper
  appends each transition to a history buffer (shown with $N{=}7$; actual implementation
  uses $N{=}17$).}
  \vspace{-2ex}
\end{figure}

\subsection{History Buffer and Zero-Phase Target Generation}

We use $N=17$ throughout, giving centered index $m=\lfloor N/2\rfloor=8$. In the
steady-state regime (buffer full), each step applies the per-dimension SG filter and
reads out the value at the middle slot:
\begin{equation}
\begin{aligned}
  \tilde{a}_{t-m}^{(d)} = \text{SG}&\left(a_{t-N+1}^{(d)},\ldots,a_t^{(d)};\,w_d,p{=}2\right)\![m], \\
  &d\in\{0,\ldots,A{-}1\}, \label{eq:sg_filter}
\end{aligned}
\end{equation}
then pushes the middle-slot transition to the replay buffer paired with $\tilde{a}_{t-m}$.
The SG window $w_d$ is specified per action dimension --- a wider window for slowly-varying
signals (steering) and a narrower one for dimensions carrying sharp transients (throttle,
speed setpoint).

Two boundary cases are handled explicitly. For the first $N$ steps after each reset, the
buffer is not yet full and we add transitions with the identity target $\tilde{a}_t=a_t$
(so the decoupled actor reproduces $\pi_\phi$'s raw action on the small fraction of
episode-startup samples). These identity-targeted transitions form a few percent of the
replay distribution; since the deployed actor has no step counter, they produce no
observable inference-time degradation at episode startup. At episode end, a terminal-flush
procedure drains the unflushed tail using one-sided polynomial fits at boundary positions;
this loses strict zero-phase at the trajectory tail but avoids underweighting terminal
states. The SG output is scaled into $[-1,1]$ and clamped to
$[-1+\varepsilon,1-\varepsilon]$ with $\varepsilon=10^{-3}$ to keep targets inside the
$\tanh$ learnable region.

\subsection{Decoupled Actor Loss with Magnitude Matching}
\label{sec:loss}

Let $\psi$ denote the parameters of the decoupled actor network $\pi_\psi$, and let
$\mu_\psi(s)$ denote its pre-squash mean output. The imitation MSE on a minibatch sampled
from the replay buffer is
\begin{equation}
  \mathcal{L}_\text{imit}(\psi) = \mathbb{E}_{(s,\tilde{a})\sim\mathcal{B}}
    \!\left[\bigl\|\tanh(\mu_\psi(s)) - \tilde{a}\bigr\|^2\right]. \label{eq:imit_loss}
\end{equation}
The total decoupled-actor loss applied at each gradient step is
\begin{equation}
  \mathcal{L}(\psi) = c \cdot \mathcal{L}_\text{imit}(\psi), \qquad
  c = \frac{\langle|\mathcal{L}_\text{actor}|\rangle}{\langle|\mathcal{L}_\text{imit}|\rangle + \varepsilon}, \label{eq:mag_match}
\end{equation}
where $\langle\cdot\rangle$ denotes a cumulative running mean over all gradient steps,
$|\mathcal{L}_\text{actor}|$ is the absolute value of the base-RL actor loss on the same
minibatch, and $\varepsilon = 10^{-6}$ is a numerical-safety constant. The scale $c$ is
treated as a constant during backpropagation. Adam absorbs constant scale factors through
per-parameter normalisation, so $c$ is mechanically redundant under adaptive optimizers
but mandatory under SGD, where unmatched gradient magnitudes (action-space MSE
$\sim 10^{-3}$ vs.\ $|Q|\sim 40$--$220$) cause the distillation to collapse. We confirm
this with a $2\times2$ optimizer ablation in \S\ref{sec:ablations}. A per-sample
Q-aware teacher selector that gates between raw and SG-filtered targets via the critic
was also considered, but proved empirically inert at every operating point we tested and
was removed (Appendix~\ref{app:qgate}).

\section{Experimental Setup}

We evaluate ZAPS-DA on two driving simulators with structurally different reward shapes.
\textbf{MetaDrive}~\cite{metadrive} (primary) is the shipped single-agent procedural
environment ($\textit{map}=3$, $\textit{traffic\_density}=0.1$,
$\textit{num\_scenarios}=1000$, $\textit{use\_lateral\_reward}=$True) at 10 Hz, with 259
scalar observation features (vehicle state + lidar) and a 2-D action (steering,
throttle/brake) in $[-1,1]$. The shipped reward is a sum of dense longitudinal-progress
$\Delta d_t \cdot \ell_t$ and speed terms with terminal $\{+10,-5,-5\}$ for
$\{$success, crash, out-of-road$\}$ --- \emph{smoothness-neutral} (no terms favouring
temporally consistent actions). Episodes terminate on arrival, crash, out-of-road, or
1000-step timeout.

\textbf{Webots ACC}~\cite{webots} (generalisation) is a custom two-vehicle car-following
task at 50 Hz: a lead vehicle follows yellow lane markings on a curved closed-loop road
with a randomised speed profile, while the ego vehicle must follow at safe distance using
a 7-dim observation (ego speed, yaw rate, previous steering and speed commands, lead
distance, lead lateral offset, lead-lost flag). The action space is steering
$\in[-0.5,0.5]$ rad and a speed setpoint $\in[0,70]$ km/h. The shipped reward sums
following-distance Gaussian, close-following penalty, lateral-alignment, yaw-comfort, and
speed-matching terms with $-100$ on crash or lead-lost --- \emph{mildly
smoothness-friendly} via Gaussian peaks favouring steady operating points. Episodes
terminate at 2100 steps ($\approx$42~s) or task failure.

All policies (main actor, decoupled actor, twin critics) are MLPs of $2\times 256$ ReLU
units; actors output $\tanh$-squashed Gaussians, critics output a scalar. Both actors are
identical in architecture but share no parameters. We add no smoothness or action-rate
penalty to either reward; ZAPS-DA's smoothness signal lives entirely outside the MDP.
The ZAPS-DA-specific configuration (SG window, polynomial order, magnitude-matching) and
network architecture are shared across both environments; standard SAC training
hyperparameters are retuned for Webots's higher control frequency and longer episodes
(Table~\ref{tab:hparams}).

\begin{table}[t]
  \centering
  \caption{Training Hyperparameters (MetaDrive anchor). Values marked $\dagger$ are
  retuned for Webots (buffer $5\times10^5$, warmup $10^4$, learning-rate $10^{-4}$);
  architecture and SG configuration are unchanged.}
  \label{tab:hparams}
  \begin{tabular}{@{}ll@{}}
    \toprule
    \textbf{Parameter} & \textbf{Value} \\
    \midrule
    SG window per action dim $(a_0, a_1)$  & $(17, 7)$ \\
    SG polynomial order $p$                & 2 \\
    SG output clamp margin $\varepsilon$   & $10^{-3}$ \\
    Learning rate (all networks + $\alpha$) & $3\times10^{-4}$\,$\dagger$ \\
    Optimizer                              & Adam (anchor) \\
    Replay buffer capacity                 & $1\times10^6$\,$\dagger$ \\
    Minibatch size                         & 512 \\
    Warmup steps                           & $70{,}000$\,$\dagger$ \\
    Discount factor $\gamma$               & 0.99 \\
    Polyak coefficient $\tau$              & 0.005 \\
    Total environment steps                & $2\times10^6$ \\
    \bottomrule
  \end{tabular}
\end{table}

\subsection{Baseline and Evaluation Protocol}
\label{sec:protocol}

\textbf{The main actor IS the SAC baseline.} ZAPS-DA's main actor is trained with the
unmodified SAC loss; given matched hyperparameters and seeds, its training trajectory is
bit-identical to a standalone vanilla SAC run. The main-actor column in every results
table is therefore vanilla SAC at matched seeds.

\textbf{Evaluation.} We use pre-registered seed bases with per-episode seeding
$\text{seed}=\text{seed\_base}+\text{episode\_index}$ so both actors see identical
scenario rollouts: MetaDrive anchor $\{0,50,150\}$ ($n=150$ paired); MetaDrive ablation
$\{0,50\}$ ($n=100$); Webots ACC $\{0,70,150\}$ ($n=150$). Inference is deterministic.
Statistics are paired $t$-tests on per-episode differences (decoupled $-$ main); the
headline MetaDrive table applies Bonferroni correction across 13 metrics
($\alpha_\text{adj}\approx 0.0038$). For each action dimension we compute four jitter
measures per episode: variance, mean absolute difference (MAD: mean of
$|a_t-a_{t-1}|$), mean second difference (MDD: mean of $|a_t-2a_{t-1}+a_{t-2}|$), and
jitter energy (JE: MSE of action minus moving average). Lower values indicate smoother
actions in all four metrics.

\textbf{Matched-seed comparison protocol.} The baseline comparison of \S\ref{sec:caps}
uses training seeds $\{0,25,125\}$ with $n=150$ deterministic test episodes per method per
seed (pooled $n=450$). Checkpoints for \emph{every} method are selected by an identical
reward-only rule: EMA plateau peaks over the post-40\%-of-training window, the raw-best
saved checkpoint within each peak neighbourhood, then a paired $n{=}25$ validation
cross-check; jitter metrics are never used for selection, so all reported jitter ratios
are genuinely held-out readouts. For these comparisons we report both pooled
($n=450$ episodes) and per-seed ($n=3$ seed means) tests --- the pooled test is powerful
but treats episodes within a seed as independent, while the per-seed test is conservative
but underpowered at $n=3$. The 15 pooled tests are reported individually uncorrected; we
therefore treat as established only effects that are pooled-significant \emph{and}
directionally consistent across all three seeds, and we explicitly flag the two cells
where the two granularities disagree (\S\ref{sec:caps}).

\section{Results}
\label{sec:results}

\subsection{MetaDrive Anchor}

\begin{figure}[t]
  \centering
  \includegraphics[width=\columnwidth]{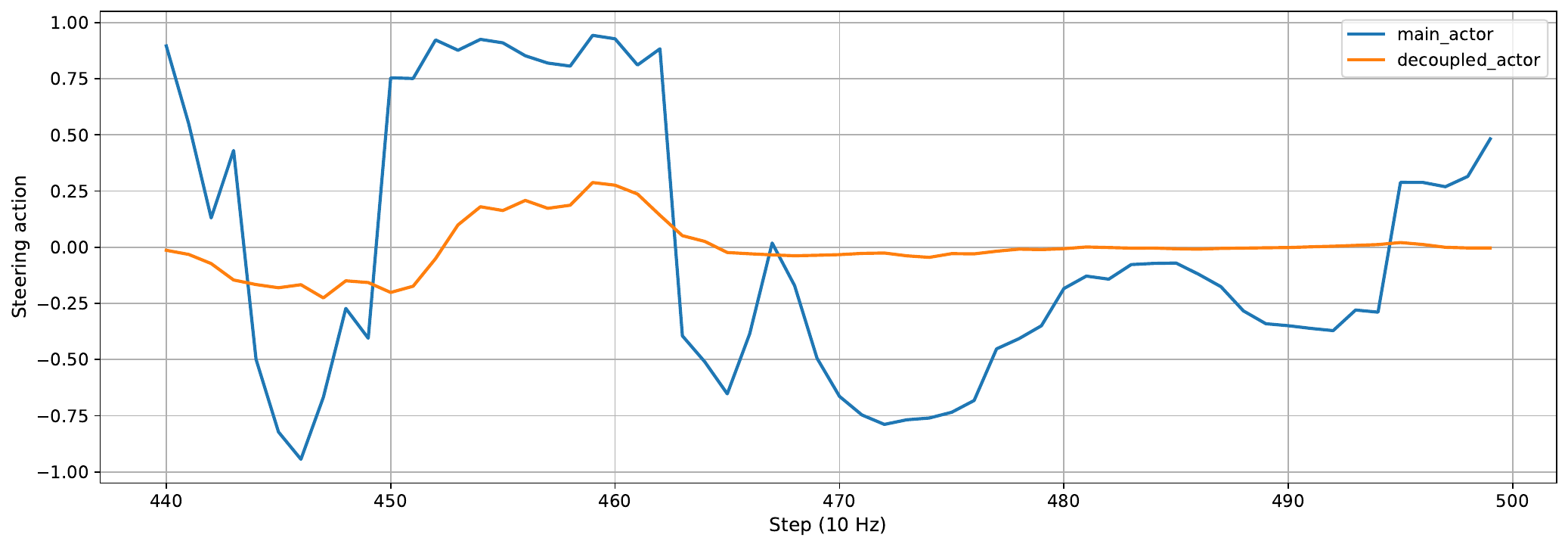}
  \caption{Steering command trace over a 60-step (6 s) window from a representative
  MetaDrive evaluation episode. The decoupled actor (orange) tracks the main actor's
  (blue) slow-varying control intent while attenuating its high-frequency oscillations.
  Reduced amplitude at sharp transitions reflects high-frequency attenuation, not
  temporal lag: on the Loop-B evaluation trace, peak cross-correlation between main
  and decoupled steering streams occurs at $+1$ step ($\approx 100$\,ms at 10\,Hz),
  versus the $(w-1)/2 = 8$-step group delay of a causal filter with the same kernel
  (Appendix~\ref{app:lag}).}
  \label{fig:steering_trace}
\end{figure}

\begin{figure}[t]
  \centering
  \includegraphics[width=\columnwidth]{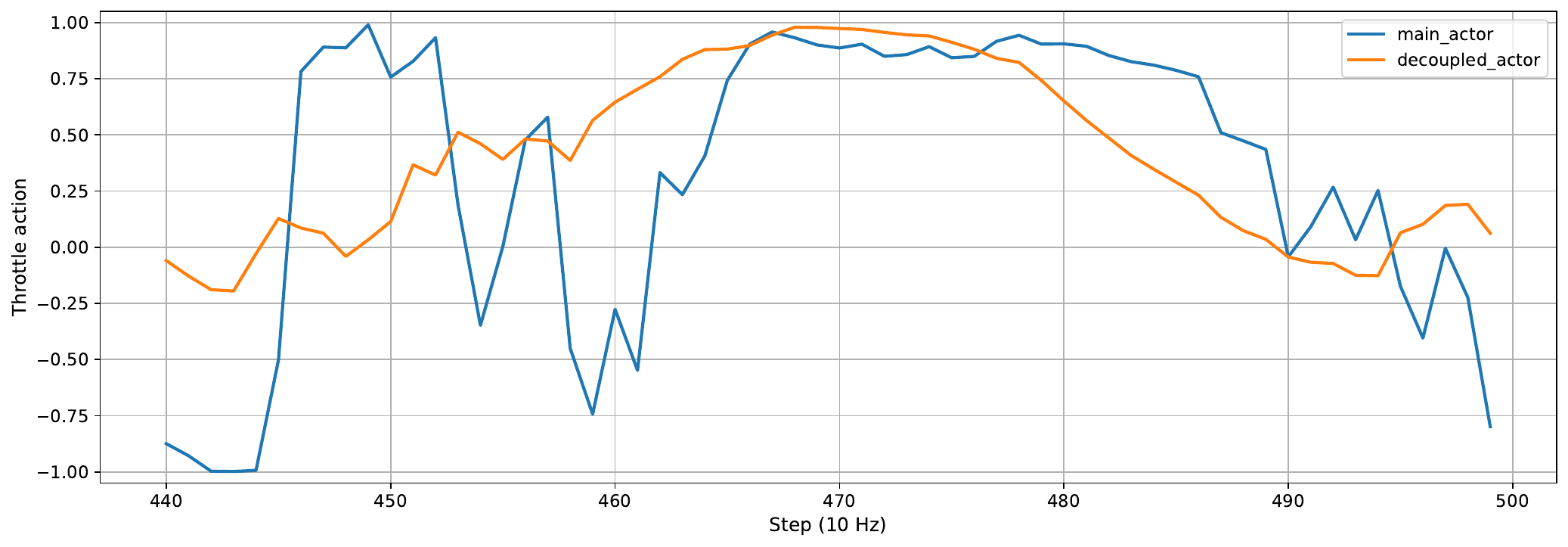}
  \caption{Throttle command trace from the same evaluation window as
  Fig.~\ref{fig:steering_trace}. The decoupled actor preserves the slow accel/cruise/decel
  envelope while removing the high-frequency reversals between full-throttle and
  full-brake. As with the steering channel, the cross-correlation between main and
  decoupled throttle streams on the Loop-B trace is indistinguishable from zero lag
  at the 100\,ms sample interval, versus the $(w-1)/2 = 3$-step group delay of a
  causal filter with the same kernel (Appendix~\ref{app:lag}).}
  \label{fig:throttle_trace}
\end{figure}

The headline result is reported at the anchor configuration --- SG window $(17,7)$,
$p=2$, 1.59M-step training checkpoint --- evaluated over $n=150$ paired episodes across
seed bases $\{0,50,150\}$. The jitter reduction is consistent across seeds (decoupled
steering JE $(0.0063, 0.0068, 0.0060)$, coefficient of variation $\approx 6\%$).
Figures~\ref{fig:steering_trace} and~\ref{fig:throttle_trace} show the qualitative
difference on each action dimension over a representative 60-step (6 s) window.

Pooled paired-test results (Table~\ref{tab:pooled_md}) show all eight jitter metrics
significantly reduced ($p<10^{-4}$, surviving Bonferroni at $\alpha_\text{adj}\approx
0.0038$); task-performance metrics (success, crash, out-of-road, length) show no
significant difference. The reward difference ($-17.8$, $-6.3\%$ relative) is marginally
significant uncorrected ($p=0.040$) but does not survive Bonferroni: it sits at the
significance boundary ($\approx 17$ reward units) and below the protocol's conventional
$80\%$-power MDE of $\approx 24$ units (Appendix~\ref{app:mde}).

\begin{table*}[t]
  \centering
  \caption{MetaDrive pooled paired-test results ($n=150$ paired episodes). Differences
  are decoupled $-$ main. Significance after Bonferroni correction for 13 comparisons.
  Ratio is main$\div$decoupled mean.}
  \label{tab:pooled_md}
  \setlength{\tabcolsep}{4pt}
  \begin{tabular}{@{}lrrrrrrr@{}}
    \toprule
    \textbf{Metric} & \textbf{Main} & \textbf{Decoupled} & \textbf{Ratio} & $\boldsymbol{\Delta\pm\text{std}}$ & $\boldsymbol{t}$ & $\boldsymbol{p}$ & \textbf{Sig.} \\
    \midrule
    Reward       & $284.34\pm94.96$  & $266.54\pm93.26$  & ---          & $-17.80\pm105.13$ & $-2.07$  & 0.040    & no \\
    Length       & $428.26\pm153.84$ & $409.19\pm150.50$ & ---          & $-19.07\pm169.95$ & $-1.37$  & 0.171    & no \\
    Arrive rate  & $0.860\pm0.347$   & $0.813\pm0.390$   & ---          & $-0.047\pm0.522$  & $-1.09$  & 0.276    & no \\
    Crash rate   & $0.113\pm0.317$   & $0.153\pm0.360$   & ---          & $+0.040\pm0.476$  & $+1.03$  & 0.305    & no \\
    Out-of-road  & $0.020\pm0.140$   & $0.033\pm0.180$   & ---          & $+0.013\pm0.231$  & $+0.71$  & 0.481    & no \\
    Steering JE  & $0.1176\pm0.0387$ & $0.0064\pm0.0047$ & \textbf{18.4$\times$} & $-0.1112\pm0.0373$ & $-36.54$ & $<10^{-4}$ & \textbf{yes} \\
    Steering MAD & $0.2556\pm0.0586$ & $0.0183\pm0.0080$ & \textbf{14.0$\times$} & $-0.2373\pm0.0597$ & $-48.71$ & $<10^{-4}$ & \textbf{yes} \\
    Steering MDD & $0.4328\pm0.1156$ & $0.0208\pm0.0090$ & \textbf{20.8$\times$} & $-0.4120\pm0.1168$ & $-43.21$ & $<10^{-4}$ & \textbf{yes} \\
    Steering Var & $0.1574\pm0.0554$ & $0.0216\pm0.0196$ & 7.3$\times$  & $-0.1358\pm0.0517$ & $-32.17$ & $<10^{-4}$ & \textbf{yes} \\
    Speed JE     & $0.0923\pm0.0287$ & $0.0238\pm0.0105$ & 3.9$\times$  & $-0.0685\pm0.0273$ & $-30.77$ & $<10^{-4}$ & \textbf{yes} \\
    Speed MAD    & $0.1633\pm0.0478$ & $0.0493\pm0.0098$ & 3.3$\times$  & $-0.1140\pm0.0491$ & $-28.43$ & $<10^{-4}$ & \textbf{yes} \\
    Speed MDD    & $0.2690\pm0.0879$ & $0.0550\pm0.0125$ & 4.9$\times$  & $-0.2139\pm0.0874$ & $-29.97$ & $<10^{-4}$ & \textbf{yes} \\
    Speed Var    & $0.2008\pm0.0678$ & $0.1362\pm0.0412$ & 1.5$\times$  & $-0.0647\pm0.0633$ & $-12.51$ & $<10^{-4}$ & \textbf{yes} \\
    \bottomrule
  \end{tabular}
\end{table*}

\begin{figure}[t]
  \centering
  \includegraphics[width=\columnwidth]{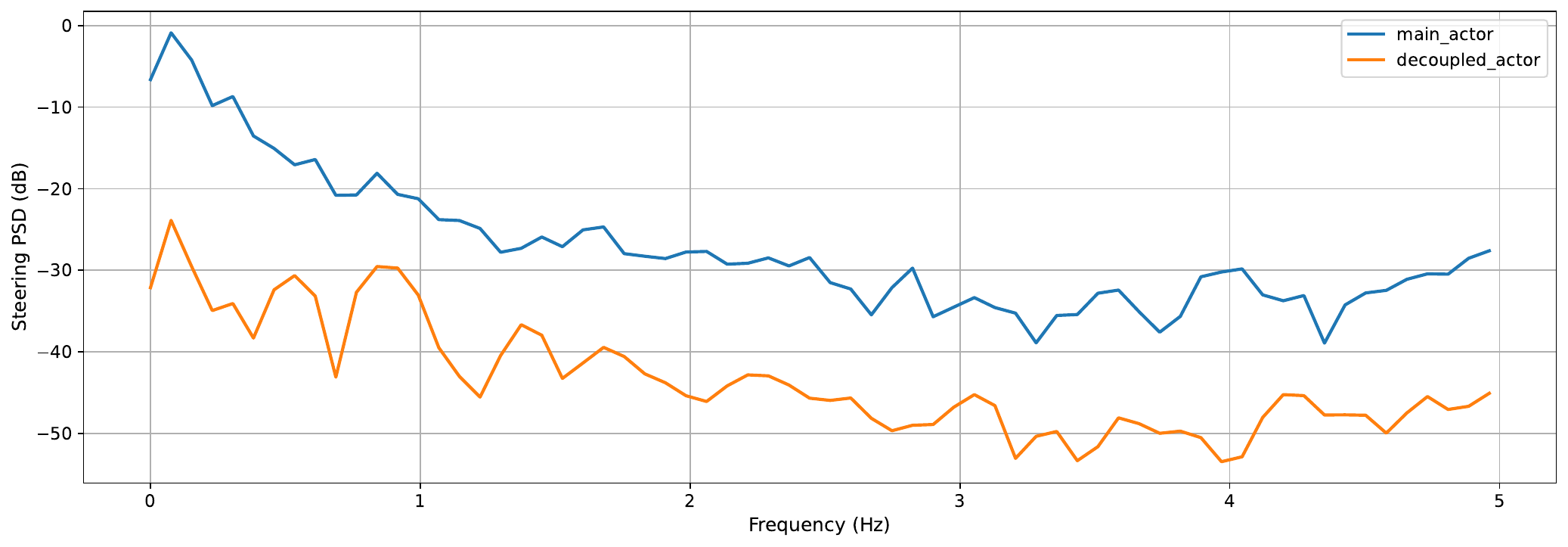}
  \caption{Power spectral density of steering action across 150 evaluation episodes.}
  \label{fig:psd}
\end{figure}

Steering jitter reduction is $14$--$21\times$ across the step-to-step metrics
(JE/MAD/MDD); throttle-axis reduction is $3$--$5\times$. Speed Var is the weakest signal
($1.5\times$) because it captures absolute throttle-command variation across an episode,
much of which reflects legitimate task-driven variation (start, brake, curve approach);
we treat MDD as the primary throttle-jitter measure. Figure~\ref{fig:psd} corroborates
in the frequency domain: the decoupled actor's steering PSD lies $10$--$20$~dB below
the main actor's across the entire band, with steeper rolloff above 2~Hz.

The task-completion profile is statistically indistinguishable (arrive $0.81$ vs.\ $0.86$
$p=0.28$; crash $0.15$ vs.\ $0.11$ $p=0.31$). The crash-rate point estimate is nonetheless
higher for the decoupled actor; this direction replicates at the matched-seed protocol of
\S\ref{sec:caps}, where the pooled difference reaches significance --- we analyse the
mechanism there and carry it into the limitations (\S\ref{sec:conclusion}). The $6.3\%$
reward cost is the trade-off for distilling the main actor's aggressive high-frequency
steering corrections that chase fine-grained gradients in MetaDrive's dense reward shape;
this cost is controlled by the SG window size (\S\ref{sec:ablations}).

\subsection{Generalisation to Webots ACC}

To test whether ZAPS-DA's smoothness mechanism transfers across environments,
we evaluate on Webots ACC with the same network sizes, batch size, SG window,
polynomial order, and magnitude-matching scheme; standard SAC hyperparameters
(learning rate, replay buffer, warmup) are retuned for Webots's 50~Hz control
frequency and longer episodes (Table~\ref{tab:hparams}). The Webots ACC result is at
the 745.5K-step checkpoint over $n=150$ paired episodes across seed bases
$\{0,70,150\}$. Pooled results (Table~\ref{tab:pooled_wb}): the decoupled actor matches
the main actor on cumulative reward ($\Delta = +49.1$, $+0.9\%$, $p=0.121$), reduces
steering jitter by $8$--$45\times$ and throttle jitter by $2.7$--$3.8\times$ (all
$p<10^{-4}$), and reduces total task-failure rate from $2.0\%$ to $0.7\%$ ---
3 lead-lost cases for the main actor, 1 ego-crash for the decoupled actor.

\begin{table*}[t]
  \centering
  \caption{Webots ACC pooled paired-test results ($n=150$). Differences are decoupled $-$ main.}
  \label{tab:pooled_wb}
  \setlength{\tabcolsep}{4pt}
  \begin{tabular}{@{}lrrrrrr@{}}
    \toprule
    \textbf{Metric} & \textbf{Main} & \textbf{Decoupled} & \textbf{Ratio} & $\boldsymbol{\Delta}$ & $\boldsymbol{t}$ & $\boldsymbol{p}$ \\
    \midrule
    Reward       & 5390.5 & 5439.7 & 1.01$\times$ & $+49.1$ (+0.9\%) & $+1.56$ & 0.121 \\
    Length (steps) & 2091.4 & 2091.8 & 1.00$\times$ & $+0.4$ & $+0.04$ & 0.971 \\
    Steering JE  & 0.0922 & 0.0021 & \textbf{44.8$\times$} & $-0.0901$ & $-485.8$ & $<10^{-4}$ \\
    Steering MAD & 0.2488 & 0.0312 & \textbf{8.0$\times$}  & $-0.2176$ & $-450.0$ & $<10^{-4}$ \\
    Steering MDD & 0.3598 & 0.0444 & \textbf{8.1$\times$}  & $-0.3154$ & $-419.6$ & $<10^{-4}$ \\
    Steering Var & 0.1008 & 0.0059 & \textbf{17.1$\times$} & $-0.0949$ & $-620.8$ & $<10^{-4}$ \\
    Speed JE     & 0.0373 & 0.0099 & 3.8$\times$ & $-0.0274$ & $-220.3$ & $<10^{-4}$ \\
    Speed MAD    & 0.1634 & 0.0615 & 2.7$\times$ & $-0.1019$ & $-284.6$ & $<10^{-4}$ \\
    Speed MDD    & 0.2375 & 0.0719 & 3.3$\times$ & $-0.1656$ & $-253.3$ & $<10^{-4}$ \\
    Speed Var    & 0.0454 & 0.0132 & 3.4$\times$ & $-0.0322$ & $-242.3$ & $<10^{-4}$ \\
    \bottomrule
  \end{tabular}
\end{table*}

The Webots result contrasts qualitatively with MetaDrive on the reward axis: a small
cost ($-6.3\%$) on MetaDrive's smoothness-neutral reward, parity ($+0.9\%$) on Webots's
mildly smoothness-friendly reward whose Gaussian shapes on lead-distance, lateral, and
speed-matching peak when the controller maintains steady operating points. The same architecture, filter, and magnitude-matching scheme produce a controlled Pareto
point on MetaDrive and a Pareto improvement on Webots --- the sign of the reward effect
depends on the environment's reward shape, not on any ZAPS-DA-side change.

\subsection{Comparison to CAPS and Post-hoc Filtering}
\label{sec:caps}

To position ZAPS-DA against the standard competing approach, we implement
CAPS~\cite{caps} --- the penalty-based smoother --- on the identical MetaDrive
environment, network, and SAC configuration. To avoid understating CAPS, we evaluate it
in \emph{both} of its operating modes: (a)~\textbf{auto-entropy}, integrated into the
auto-entropy SAC used throughout this work, changing only the actor loss (the two CAPS
penalty terms on the $\tanh$-squashed mean action, so the policy is otherwise
bit-identical to our vanilla-SAC baseline); and (b)~\textbf{native fixed-entropy}, the
form in which CAPS was published, additionally changing the entropy scheme to a fixed
coefficient --- re-tuned to $\alpha=0.02$, because the published $\alpha=0.2$ prevents
even vanilla SAC from learning on MetaDrive (finding~iii). In each mode we sweep the
penalty weight $\lambda_T$ (canonical ratio $\lambda_S=5\lambda_T$; spatial-noise scale
$\sigma=0.05$, calibrated to MetaDrive's normalized observations) and select checkpoints
with the reward-only rule of \S\ref{sec:protocol} applied identically to every method. This is an
independently controlled comparison at the matched-seed protocol (seeds $\{0,25,125\}$,
$n=150$ paired test episodes per seed, 2M steps); absolute values therefore differ from
the anchor of Table~\ref{tab:pooled_md} (ZAPS-DA steering-JE reduction is $14.7\times$
here vs.\ the anchor's $18.4\times$).

\begin{table}[t]
  \centering
  \caption{CAPS vs.\ ZAPS-DA on MetaDrive at the matched-seed protocol (seeds
  $\{0,25,125\}$, $n=150$ test episodes per seed, pooled $n=450$, 2M steps; mean $\pm$
  cross-seed std). Jitter ratios are vanilla$\div$method against the common auto-entropy
  vanilla anchor; against its own fixed-$\alpha$ control, CAPS-native steering-JE
  reduction is $2.7\times$.}
  \label{tab:caps}
  \setlength{\tabcolsep}{2pt}
  \footnotesize
  \resizebox{\columnwidth}{!}{%
  \begin{tabular}{@{}lrrrrr@{}}
    \toprule
    \textbf{Method} & \textbf{Reward} & \textbf{Arrive} & \textbf{Crash} & \textbf{JE$\downarrow$} & \textbf{MAD$\downarrow$} \\
    \midrule
    Vanilla SAC                            & $252\pm4$  & $.83\pm.04$ & $.09\pm.02$ & $1.0\times$ & $1.0\times$ \\
    CAPS auto-$\alpha$ ($\lambda_T{=}0.15$)      & $292\pm6$  & $.93\pm.01$ & $.07\pm.02$ & $2.0\pm0.5\times$ & $1.9\pm0.4\times$ \\
    CAPS fix.-$\alpha{=}.02$ ($\lambda_T{=}0.5$) & $282\pm12$ & $.86\pm.05$ & $.12\pm.04$ & $3.2\pm0.6\times$ & $2.7\pm0.2\times$ \\
    ZAPS-DA                                & $251\pm1$  & $.80\pm.01$ & $.14\pm.04$ & $\mathbf{14.7\pm3.4\times}$ & $\mathbf{11.3\pm1.5\times}$ \\
    \bottomrule
  \end{tabular}}
\end{table}

Three findings (Table~\ref{tab:caps}). \textbf{(i) Smoothness ceiling.} Even at its best
native operating point --- fixed entropy coefficient re-tuned to the environment,
re-centered penalty weight, cross-seeded, and checkpoint-selected by the shared rule ---
CAPS reaches $3.2\times$ steering-jitter reduction (per-seed $\{2.6, 3.0, 3.8\}\times$),
against ZAPS-DA's $14.7\times$ at the same protocol: a $\approx 4.6\times$ gap. The
native fixed-$\alpha$ form is smoother than the auto-entropy form ($3.2\times$ vs.\
$2.0\times$), confirming that CAPS is represented here at its genuine best rather than
handicapped by a mis-integration --- and it still falls several-fold short.

\textbf{(ii) Task--smoothness positions.} The two CAPS variants occupy a gentler,
higher-arriving region of the trade-off, while ZAPS-DA trades a small task cost for far
larger smoothing. Following our two-granularity reporting standard: CAPS-native's reward
advantage over ZAPS-DA ($+31$) is pooled-significant but borderline per-seed ($p=0.055$);
its arrive advantage is pooled-significant but not per-seed; crash differences between
CAPS-native and ZAPS-DA are not significant at either granularity. CAPS-auto's arrive
gain over vanilla is likewise pooled-significant only (per-seed $p=0.053$), and its crash
improvement is non-significant --- so we characterise CAPS-auto as a gentler,
higher-arriving policy, not as a safer one. For ZAPS-DA vs.\ vanilla, task-completion is
at parity (McNemar pooled $p=0.30$, per-seed $p=0.54$), but the crash rate is
directionally higher in all three seeds ($+0.093/{+}0.033/{+}0.027$; pooled McNemar
$p=0.013$, per-seed $p=0.137$). These are the two cells where pooled and per-seed tests
disagree (\S\ref{sec:protocol}); we treat the CAPS-auto arrive gain as suggestive, and we treat the
ZAPS-DA crash increase as real --- its mechanism is the one exposed by the
symmetric-window ablation (\S\ref{sec:ablations}): filtering the throttle axis attenuates
part of the sharp reactive-braking transients the policy uses near traffic. We return to
this in the limitations (\S\ref{sec:conclusion}).

\textbf{(iii) No published CAPS configuration transfers; the tuning cascade is the cost.}
Under the auto-entropy SAC standard in this work, CAPS's published weight
($\lambda_T=20$) diverges: the smoothness penalty suppresses policy entropy, SAC's
automatic temperature tuner cranks it back up, and the coefficient runs away to numerical
overflow --- forcing a usable weight range roughly two orders of magnitude below the
published value, itself seed-fragile near its ceiling (the smoothest stable seed-0 point
diverges at another seed). In the published fixed-entropy form, the released coefficient
($\alpha=0.2$) is mis-scaled for MetaDrive's reward magnitude and prevents even vanilla
SAC from learning; the environment's usable fixed-$\alpha$ window is narrow
($\alpha\approx 0.02$; $\alpha=0.05$ already collapses). With $\alpha$ fixed at the
working value, higher $\lambda$ trades divergence for collapse into degenerate
near-constant actions, and late-training dynamics are seed-sensitive, requiring careful
early checkpoint selection. Reproducing CAPS on this higher-dimensional driving task thus
required re-tuning the spatial-noise scale, the penalty weight ($\sim\!10^2\times$ off
published), \emph{and} the entropy coefficient, plus per-seed stability handling.
ZAPS-DA, whose smoothness signal never enters the RL gradient, introduces no
RL-coupled hyperparameters and is seed-robust; its only knobs are the SG window sizes ---
an interpretable smoothness--reward Pareto knob (\S\ref{sec:ablations}) that transferred
to Webots unchanged.

\textbf{Post-hoc filtering trade-off, and where each edge lies.} Applying a causal
moving-average filter offline to the vanilla action stream recovers smoothness only at
the cost of phase lag: matching ZAPS-DA's steering-jitter reduction requires a window
whose group delay is $\approx 0.84$\,s at 10\,Hz. The comparisons are thus complementary:
against the causal filter, ZAPS-DA's edge is \emph{zero lag} at matched smoothness;
against CAPS --- which, like ZAPS-DA, deploys a zero-lag feed-forward policy --- the edge
is instead \emph{magnitude} ($\approx 4.6\times$ more jitter reduction), \emph{zero
per-environment tuning} of the smoothness signal, and \emph{post-hoc applicability}:
ZAPS-DA can be attached to an already-trained policy, which no penalty-based method
affords.

\section{Ablations}
\label{sec:ablations}

\subsection{SG Window Sensitivity}

Five training runs sweep the steering window over $\{9, 13, 17$ (anchor)$, 21\}$ with
the throttle window fixed at 7, plus one symmetric run $(17,17)$ that widens the throttle
window to match steering. All runs share identical SAC hyperparameters and use $p=2$;
all jitter-reduction differences are significant at $p<10^{-4}$ in every condition.
Table~\ref{tab:sg_ablation} summarises the runs.

\begin{table*}[t]
  \centering
  \caption{SG window sensitivity ablation (paired $t$-test on decoupled $-$ main, per-condition $n$ as noted).}
  \label{tab:sg_ablation}
  \setlength{\tabcolsep}{4pt}
  \begin{tabular}{@{}llrrrrrrrrr@{}}
    \toprule
    \textbf{Run} & \textbf{Window} & $\boldsymbol{n}$ & \textbf{Main} & \textbf{Decoupled} & \textbf{\% cost} & $\boldsymbol{p}$ & \textbf{Arrive (m/d)} & \textbf{Crash (m/d)} & \textbf{JE ratio} & \textbf{MDD ratio} \\
    \midrule
    W1 & $(9, 7)$            & 100 & 272.0 & 265.8 & $-2.3\%$  & 0.443    & 0.78 / \textbf{0.89} & 0.09 / 0.08 & 10.9$\times$ & 3.5$\times$ \\
    W2 & $(13, 7)$           & 100 & 304.3 & 287.0 & $-5.7\%$  & 0.057    & 0.91 / 0.83 & 0.08 / 0.15 & 14.9$\times$ & 5.2$\times$ \\
    \textbf{W3} & \textbf{(17,7) anchor} & \textbf{150} & \textbf{284.3} & \textbf{266.5} & \textbf{$-6.3\%$} & \textbf{0.040} & \textbf{0.86 / 0.81} & \textbf{0.11 / 0.15} & \textbf{18.4$\times$} & \textbf{4.9$\times$} \\
    W4 & $(21, 7)$           & 100 & 303.2 & 265.4 & $-12.5\%$ & $<10^{-4}$ & 0.91 / 0.81 & 0.08 / 0.16 & 15.5$\times$ & 5.5$\times$ \\
    W5 & $(17, 17)$ symm.    & 100 & 277.5 & \textbf{212.3} & \textbf{$-23.5\%$} & $<10^{-4}$ & 0.84 / \textbf{0.55} & 0.13 / \textbf{0.39} & 14.3$\times$ & 4.2$\times$ \\
    \bottomrule
  \end{tabular}
\end{table*}

\textbf{The SG window is the smoothness--reward Pareto knob.} Reward cost grows
monotonically with steering window size on the throttle-fixed sweep:
W1 ($-2.3\%$) $\to$ W2 ($-5.7\%$) $\to$ W3 ($-6.3\%$) $\to$ W4 ($-12.5\%$). However,
steering jitter reduction is \emph{non-monotonic}: W1 ($10.9\times$) $\to$ W2
($14.9\times$) $\to$ W3 ($18.4\times$) $\to$ W4 ($15.5\times$). W4 is a strict Pareto
regression relative to W3 --- more reward cost, less jitter reduction --- because longer
SG windows have more aggressive negative side-lobes that re-introduce ripple at
frequencies the shorter window suppresses. The anchor $(17, 7)$ therefore sits at the
empirical operating-point elbow.

\textbf{The per-dimension narrow-throttle window is load-bearing.} The symmetric
ablation W5 widens the throttle window from 7 to 17, matching steering, and produces the
worst run in the sweep: decoupled reward drops $20\%$, arrive rate falls from 0.81 to
0.55, and crash rate nearly triples from 0.15 to 0.39. The mechanism is direct: a
17-wide throttle filter averages out the brake/accel transients that the policy uses
for reactive deceleration, causing the decoupled actor to under-brake into preceding
traffic. The narrow throttle window is therefore not stylistic --- it is necessary to
preserve task safety on MetaDrive, and provides the clearest evidence that the
SG-filter teacher attenuates exactly the action content the policy uses for reactive
control.

\subsection{Magnitude-Matching Mechanism}

A $2\times2$ factorial ablation crosses decoupled-actor optimizer
$\in\{\text{Adam}, \text{SGD w/ momentum 0.9}\}$ with magnitude-matching
$\in\{\text{with-scale}, \text{without-scale}\}$, all at the anchor configuration
(Table~\ref{tab:mag_match}).

\begin{table*}[t]
  \centering
  \caption{Magnitude-matching ablation (paired $t$-test on decoupled $-$ main).}
  \label{tab:mag_match}
  \setlength{\tabcolsep}{4pt}
  \begin{tabular}{@{}lrrrrrrrrr@{}}
    \toprule
    \textbf{Condition} & $\boldsymbol{n}$ & \textbf{Main} & \textbf{Decoupled} & $\boldsymbol{\Delta}$ & \textbf{\% cost} & $\boldsymbol{p}$ & \textbf{Arrive (m/d)} & \textbf{Crash (m/d)} & \textbf{JE ratio} \\
    \midrule
    Adam + scale (anchor) & 150 & 284.3 & 266.5 & $-17.8$ & $-6.3\%$  & 0.040    & 0.86 / 0.81 & 0.11 / 0.15 & 18.4$\times$ \\
    Adam $-$ scale         & 100 & 247.2 & 244.3 & $-3.0$  & $-1.2\%$  & 0.788    & 0.67 / 0.75 & 0.06 / 0.17 & 19.0$\times$ \\
    SGD + scale            & 100 & 267.9 & 258.1 & $-9.8$  & $-3.7\%$  & 0.391    & 0.80 / 0.81 & 0.11 / 0.09 & 12.3$\times$ \\
    SGD $-$ scale          & 100 & 286.0 & \textbf{175.6} & $-110.5$ & \textbf{$-38.6\%$} & $<10^{-4}$ & 0.84 / \textbf{0.38} & 0.11 / \textbf{0.44} & 18.7$\times$ \\
    \bottomrule
  \end{tabular}
\end{table*}

Adam $+$ scale and Adam $-$ scale produce reward-comparable decoupled actors with
near-identical jitter ratios ($18.4\times$ vs.\ $19.0\times$ steering JE), confirming
Adam's empirical scale-invariance through per-parameter normalisation: the
magnitude-matching scale is mechanically redundant under adaptive optimizers. Under SGD
the picture changes sharply. SGD $+$ scale produces a decoupled actor reward-tied with
the main actor ($\Delta = -9.8$, $p=0.391$), comparable to Adam in safety and reward.
SGD $-$ scale collapses catastrophically: reward drops to 175.6 ($-38.6\%$ paired
cost), arrive rate falls to 0.38, crash rate rises to 0.44; the collapse is reproduced
consistently across both seeds. The nominally high jitter ratio under SGD $-$ scale
($28.5\times$ steering MDD) is not a positive result but a tell of the failure mode:
without proper gradient signal, the network collapses toward a low-variance
near-constant output that fails to drive. Magnitude-matching is therefore an
optimizer-class portability device --- redundant under Adam, mandatory under SGD ---
that lets the dual-actor framework be deployed under either optimizer class at zero
additional hyperparameter cost.

\section{Conclusion and Limitations}
\label{sec:conclusion}
We presented ZAPS-DA, a framework for reducing action jitter in off-policy continuous
control through architectural decoupling: the main actor is trained with the unmodified
base RL loss, while a separate decoupled actor is trained via supervised regression
against zero-phase filtered targets stored in the replay buffer. The deployed
artifact is the decoupled actor --- a feed-forward map from current observation to
smooth action with no inference-time filter, no action-history input, and negligible phase lag.
Beyond the specific results, ZAPS-DA demonstrates a general principle: \emph{non-causal
oracles can be distilled into causal policies during training without requiring the
oracle at deployment}. The same architectural pattern admits other zero-phase operators
(Butterworth via \texttt{filtfilt}, learned smoothing kernels) and other non-causal
oracles (offline optimal controllers, look-ahead planners) wherever the oracle is
well-defined as a function of a windowed history and the training algorithm maintains
a replay buffer or analog. \textbf{Limitations.} First, scope: validated only on SAC and on autonomous
driving tasks; SG filter only; simulation-only --- the MetaDrive and Webots environments
capture the structural pathology (low-pass plant dynamics masking jitter from the critic)
that motivates ZAPS-DA, but do not model actuator noise, sensor latency, or wear-induced
parameter drift. Accordingly, the deployment-oriented claims (zero phase lag, no
post-processing) are established in simulation under this structural model and remain to
be confirmed on physical hardware. Second, a behavioural limitation we consider real: at
the matched-seed protocol (\S\ref{sec:caps}) the decoupled actor's MetaDrive crash rate
is directionally higher than vanilla SAC's in all three seeds
($+0.093/{+}0.033/{+}0.027$; pooled McNemar $p=0.013$, per-seed $p=0.137$), while
task-completion is at parity. The mechanism is the one the symmetric-window ablation
isolates: throttle-axis filtering attenuates part of the reactive-braking transients the
policy uses near traffic, and even the anchor's deliberately narrow throttle window
retains a residue of this effect. Practitioners should treat the throttle window as a
safety-relevant parameter and verify crash statistics at deployment scale before use.
Natural extensions include TD3/DDPG validation, locomotion and manipulation benchmarks,
alternative zero-phase filters, and real-hardware deployment.

\textbf{Code and data availability.} Code, checkpoints, per-seed evaluation CSVs
--- including the seeded ZAPS-DA evaluations at training seeds $\{0,25,125\}$ that
reproduce the ZAPS-DA side of Table~\ref{tab:caps} --- ablation details, failed
variants, and full reproducibility scripts are available at
\href{https://github.com/fshamass/ZAPS-DA}{github.com/fshamass/ZAPS-DA}.

\appendices

\section{Failed Variants}
\label{app:failed}

\subsection{Lagrangian merged-objective penalty}
\label{app:lagrangian}
The natural way to add smoothness to off-policy RL is to append the SG-imitation MSE
directly to the SAC actor loss under a dual-tuned coefficient $\lambda$:
\begin{align*}
  \mathcal{L}_\pi=\mathbb{E}\bigl[
    &\alpha\log\pi(a|s)-\min_i Q_{\theta_i}(s,a) \\
    &+\lambda\,|\bar{Q}|\,\bigl\|\tanh(\mu(s))-\tilde{a}\bigr\|^2\bigr],
\end{align*}
where $\tilde a$ is the zero-phase SG target and $\lambda$ is auto-tuned by dual
ascent~\cite{pid_lag} toward an imitation-MSE target of $0.01$, hard-clamped to $[0.1,1.0]$
to prevent runaway. In a 770k-step MetaDrive run at otherwise-anchor configuration,
$\lambda$ \emph{saturated at its ceiling of $1.0$ for the entire run}; the imitation MSE
plateaued at $\approx 0.024$ (never reaching $0.01$); and the main actor's steering JE stayed
at vanilla-SAC levels ($\approx 0.112$), roughly $22\times$ the decoupled actor's
$\approx 0.005$ at the same stage. The bounded regime is structurally inert --- once
$\lambda$ reaches its ceiling the SAC objective dominates and the penalty cannot strengthen.
Removing the ceiling instead lets dual ascent drive $\lambda\!\to\!\infty$, where the
imitation term overwhelms the $(\alpha\log\pi-\min_i Q_{\theta_i})$ gradient and collapses
reward learning. This \emph{two-horned} failure --- inert when bounded, destructive when
unbounded --- is the merged-objective confound (\S\ref{sec:related}) that motivates
architectural decoupling: no single-actor operating point reduces jitter and preserves reward
together. The v1 checkpoints were not preserved, so paired re-evaluation at the main paper's
sample sizes is unavailable for this variant; the recorded signal is the \emph{absence} of
smoothing (main-actor jitter at vanilla levels throughout), whose $\approx 22\times$
magnitude is far outside sample-size noise.

\subsection{Q-aware teacher selector}
\label{app:qgate}
An earlier version gated, per sample and per action dimension $d$, between the SG target and
the raw action using the critic: it compared $Q(s,\mathbf{a})$ against
$Q(s,\mathbf{a}_{d\to\text{SG}})$ (dimension $d$ alone smoothed) and fell back to raw whenever
smoothing $d$ exceeded a per-dimension budget $\beta_d\,|Q(s,\mathbf{a})|$, otherwise adopting
the SG value. This costs $A{+}1$ critic passes per gradient step ($A$ the action
dimensionality) and adds a length-$A$ hyperparameter $\beta$. We swept $\beta$ over five
scales of the reference $[0.1,0.03]$ (2M steps each, anchor SG window; best checkpoint by
decoupled reward over the final 300k), reporting \emph{accept}, the fraction of training
samples for which the gate kept the SG target (Table~\ref{tab:qgate_sweep}).

\begin{table}[t]
  \centering
  \caption{Q-gate $\beta$-sensitivity sweep (2M steps; pooled paired evaluation, $n$ as
  noted). \emph{accept} = fraction of training samples the gate kept the SG target
  (steering/throttle). JE$\downarrow$ is main$\div$decoupled steering-JE.}
  \label{tab:qgate_sweep}
  \setlength{\tabcolsep}{3pt}
  \footnotesize
  \begin{tabular}{@{}lrrrrrr@{}}
    \toprule
    $\beta$ (scale) & $n$ & Main & Dec. & \%cost & JE$\downarrow$ & accept (st/th) \\
    \midrule
    $\beta_1$ (0.2$\times$)         & 100 & 306.8 & 287.6 & $-6.2$ & 12.1$\times$ & .99/.98 \\
    $\beta_2$ (0.5$\times$)         & 100 & 258.7 & 271.9 & $+5.1$ & 15.5$\times$ & 1.00/.99 \\
    $\beta_3$ (1$\times$, anchor)   & 150 & 284.3 & 266.5 & $-6.3$ & 18.4$\times$ & 1.00/1.00 \\
    $\beta_4$ (2$\times$)           & 100 & 306.7 & 279.9 & $-8.7$ & 11.4$\times$ & 1.00/1.00 \\
    $\beta_5$ (4$\times$)$^\dagger$ & 100 & 198.6$^\dagger$ & 256.8 & --- & 18.9$\times$ & 1.00/1.00 \\
    \bottomrule
  \end{tabular}
  \\[2pt]
  {\footnotesize $^\dagger$ $\beta_5$ main-actor reward is a single-seed SAC training failure
  ($198.6$ vs.\ $258$--$307$ elsewhere), a broken-main artifact, not a $\beta$ effect; its
  \%cost is omitted.}
\end{table}

The gate binds only at the smallest budget ($\beta_1$: accept $0.989/0.976$ for
steering/throttle); from $\beta_3$ upward it never rejects SG (accept $\approx 1.00$), so the
decoupled actor receives the SG target on essentially every sample regardless of $\beta$, and
no monotone reward-vs-jitter Pareto curve emerges --- single-seed SAC variance (main-actor
reward $258$--$307$ across identical recipes) dominates the residuals. To isolate the gate
from any $\beta$ confound we compared gate-on against gate-off (always-SG) at $\beta_1$, its
only binding setting (Table~\ref{tab:qgate_onoff}): the two are statistically
indistinguishable on reward, arrive, crash, and jitter, because a rejected action is by
construction within $\beta_1\,|\bar Q|$ of the SG target and its effect averages out over
training.

\begin{table}[t]
  \centering
  \caption{Gate-on vs.\ gate-off (always-SG) at $\beta_1$, the only binding budget
  ($n=100$ paired). m/d = main/decoupled.}
  \label{tab:qgate_onoff}
  \setlength{\tabcolsep}{2.5pt}
  \footnotesize
  \resizebox{\columnwidth}{!}{%
  \begin{tabular}{@{}lrrrrrr@{}}
    \toprule
    Condition & Main & Dec. & $\Delta r$ & Arrive (m/d) & Crash (m/d) & JE$\downarrow$ \\
    \midrule
    Gate-on ($\beta_1$)  & 306.8 & 287.6 & $-19.2$ & .89/.85 & .10/.12 & 12.1$\times$ \\
    Gate-off (always SG) & 304.1 & 284.6 & $-19.5$ & .88/.84 & .07/.15 & 12.7$\times$ \\
    \bottomrule
  \end{tabular}}
\end{table}

The gate was therefore removed for simplicity --- one fewer (vector) hyperparameter and
$A{+}1$ fewer critic passes per step --- at no measurable cost. The SG-window
(Table~\ref{tab:sg_ablation}) and magnitude-matching (Table~\ref{tab:mag_match}) results were
collected with the gate present at $\beta_3$, where it is inert and thus equivalent to
gate-free training.

\section{Evaluation Details}
\label{app:eval}

\subsection{Minimum detectable effect of the anchor protocol}
\label{app:mde}
We quantify the sensitivity of the $n=150$ paired anchor protocol on episode reward.
Across the 150 paired episodes the per-episode reward difference (decoupled $-$ main) has
standard deviation $\mathrm{SD}=105.1$, so the standard error of the mean difference is
$\mathrm{SE}=105.1/\sqrt{150}=8.58$ reward units. For a two-sided paired $t$-test at
$\alpha=0.05$ ($\mathrm{df}=149$), the minimum effect detectable at conventional power
$1-\beta=0.80$ is
\begin{align*}
  \mathrm{MDE}&=\bigl(t_{0.975,149}+t_{0.80,149}\bigr)\,\mathrm{SE} \\
              &=(1.976+0.844)\times 8.58\approx 24.2\ \text{reward units.}
\end{align*}
The observed difference of $-17.8$ (Cohen's $d_z=-0.17$) is \emph{below} this threshold:
it corresponds to an achieved power of only $\approx 0.54$ and was detected at $p=0.040$
only marginally (the $17.0$-unit effect at the $50\%$-power significance boundary is
$t_{0.975,149}\,\mathrm{SE}$). The anchor is thus powered to resolve reward differences of
$\approx 24$ units or larger; the observed $\approx 18$-unit gap sits below that, which is
why --- despite an uncorrected $p=0.040$ --- it is a fragile effect that does not survive
Bonferroni correction and is best read as a small, near-negligible reward cost.

\subsection{Lag analysis of the deployed decoupled actor}
\label{app:lag}
We quantify the phase lag of the \emph{deployed} decoupled actor by cross-correlating its
steering and throttle command streams against the main actor's on the paired Loop-B
evaluation, in which both actors act on identical state trajectories. On the representative
evaluation trace (Fig.~\ref{fig:steering_trace}) the steering cross-correlation peaks at a
lag of $+1$ to $+2$ steps ($\le 200$\,ms at 10\,Hz; the peak is broad, $r\approx0.60$ across
lags $+1$ and $+2$). Aggregated over all 150 evaluation episodes the per-episode steering lag
has median $0$ and lies within $\pm1$ step in roughly half of episodes --- the decoupled
actor is effectively zero-phase. The throttle channel is near-saturated (command
$\approx 0.95$--$0.99$ over most of the trace) and carries little dynamic range; its
cross-correlation is flat within $\pm1$ step of zero lag and admits no resolvable delay.
Both channels are far below the $(w{-}1)/2$ group delay that a \emph{causal} filter with the
same kernel incurs --- $8$ steps for the steering window ($w=17$) and $3$ steps for the
throttle window ($w=7$). The decoupled actor thus recovers the zero-phase SG target's
smoothing without the group delay a deployable causal filter would add.

\begin{IEEEbiographynophoto}{Faiq Shamass}
received the B.S.\ degree in civil engineering from the University of Baghdad,
Baghdad, Iraq, in 1989, and the M.S.\ degree in electronics and computer control
systems from Wayne State University, Detroit, MI, USA, in 2000.
He has over two decades of experience as a software engineer in the automotive,
semiconductor, and telecommunications industries, most recently as a Staff Software
Engineer at Qualcomm, Inc., San Diego, CA, USA (2013--2024), where he worked on
cellular-vehicle-to-everything (C-V2X) telematics SDKs and SystemC/TLM hardware
modeling. Since 2024 he has been an independent research engineer developing
end-to-end reinforcement-learning pipelines for autonomous-vehicle control,
including simulation-to-real transfer to embedded targets. His research interests
include reinforcement learning for continuous control, autonomous driving, and
signal processing for learned control policies.
\end{IEEEbiographynophoto}

\end{document}